# Thermal-Mechanical Physics Informed Deep Learning For Fast Prediction of Thermal Stress Evolution in Laser Metal Deposition

R. Sharma[a,b], Y.B. Guo[a,b*]

[a] Dept. of Mechanical and Aerospace Engineering, Rutgers University-New Brunswick, Piscataway, NJ 08854, USA
[b] New Jersey Advanced Manufacturing Initiative, Rutgers University-New Brunswick, Piscataway, NJ 08854, USA

*Corresponding author: yuebin.guo@rutgers.edu (Y.B. Guo)

**Abstract**

Understanding thermal stress evolution in metal additive manufacturing (AM) is crucial for producing high-quality components. Recent advancements in machine learning (ML) have shown great potential for modeling complex multiphysics problems in metal AM. While physics-based simulations face the challenge of high computational costs, conventional data-driven ML models require large, labeled training datasets to achieve accurate predictions. Unfortunately, generating large datasets for ML model training through time-consuming experiments or high-fidelity simulations is highly expensive in metal AM. To address these challenges, this study introduces a physics-informed neural network (PINN) framework that incorporates governing physical laws into deep neural networks (NNs) to predict temperature and thermal stress evolution during the laser metal deposition (LMD) process. The study also discusses the enhanced accuracy and efficiency of the PINN model when supplemented with small simulation data. Furthermore, it highlights the PINN transferability, enabling fast predictions with a set of new process parameters using a pre-trained PINN model as an online soft sensor, significantly reducing computation time compared to physics-based numerical models while maintaining accuracy.

*Keywords:* Metal additive manufacturing, thermal stress, physics-informed machine learning, transfer learning, soft sensor.

## 1. Introduction

Metal additive manufacturing (AM), including powder bed fusion (PBF) and directed energy deposition (DED), is an enabling technology for producing complex metal parts in a layer fashion, which has wide applications in the aerospace, biomedical, automotive, and energy industries. In PBF, a high-energy beam, such as a laser or electron beam, is used to completely melt thin layers of metal powder (30-50 µm thick) in an inert environment to print a part. DED creates parts by directly melting feedstock materials (e.g., powder or wire) using a laser or electron beam, which has a significantly higher material deposition rate compared to PBF (1). Despite the great potential of metal AM, its adoption in the industry remains limited due to uncertainties in the process-microstructure-property relationship. The thermal cycle in metal additive manufacturing is unique due to its rapid heating, cooling, and re-melting of previously solidified layers compared with traditional manufacturing processes, such as machining, casting, and heat treatment. The unique thermal stress cycle generates complex residual stresses, which causes a major concern (2). For example, residual stress may interfere with or damage the recoater blade during printing and deteriorate the dimension inaccuracies of a final component (3). Residual stresses may also cause fracture at lower-than-ideal applied stress levels, further weakening a component and reducing its overall functions.

Many researchers have developed numerical methods to calculate the residual stress and validated their models with destructive or non-destructive experimental methods (4–6). These methods solve the physical governing equations i.e. energy and mechanical equations by finite element analysis (FEA). Dai and Shaw (7) developed one of the earlier FEA models to calculate residual stress for laser-processed components made up of multiple materials. They modeled the part buildup by adding all the powder elements in a single layer on top of the previously deposited layer. Afterward, some of these newly added elements were subjected to laser processing, following a pre-defined scanning strategy. This process led to the creation of a complete component, while the surrounding area was left with loosely bound powder particles. It was concluded that sequence I, which builds the high melting point material first, is the best approach for parts made of two materials, while sequence II, which builds both materials layer by layer, lacks control and is unsuitable. Ding et al. (8) developed a transient 3D model to study the development of residual stress in a multi-layer build-up process. They used a series of sequential steps, where the temperature from the previous step was carried over for each newly added layer. "Element birth technique" was used to simulate the multiple layers. The developed new approach which utilizes a steady-state thermal model, offers a significant reduction in computational time. Li et al. (9) developed a geometry-scalable predictive model that spans the microscale laser scan, mesoscale layer hatch, and macroscale part build-up to efficiently predict residual stresses under various scanning strategies. These predictions were validated using L-shaped bar and bridge structures. The geometry scalability approach offers an efficient method for optimizing part designs. These conventional numerical methods for calculating the residual stress in AM involve solving PDEs by converting them into algebraic equations suitable for digital computation. While effective, these methods require complex mathematical techniques for discretization, coupling, and boundary conditions, often demanding significant computational resources and hours. Additionally, their performance varies by problem, requiring practitioners to have deep expertise in both manufacturing processes and mathematical methods.

Recently, machine learning (ML) methods have gained popularity for solving complex problems in manufacturing processes, due to the ability to capture large amounts of data through sensors (10–13). These models gave comparable accuracy compared to the conventional numerical methods. Some researchers have attempted to use deep learning (DL) models to predict the residual stress developed in AM components (14–16). These models focus on algorithmic data modeling and predicting labels based on observations, emphasizing accurate predictions for classification and regression tasks. The main advantage of ML models is their ability to transfer easily to another processing condition after being trained on a specific process condition (17). This is particularly useful for online real-time process prediction, where traditional simulation models often fail due to the high computation cost to execute, even if just one process parameter changes. However, a disadvantage of conventional ML model pre-training is their reliance on large-labeled datasets, which could be expensive to generate. Additionally, the black-box nature of conventional ML models lacks explainability, especially when the model fails to predict accurate results.

In recent years, the accuracy of ML models (especially DL) has significantly improved by leveraging available data and the governing PDEs of the process. This novel approach, known as "physics-informed neural network (PINN)" (18). This approach is superior to conventional numerical methods for the following reasons:
- PINN method is a mesh-free technique that makes the significant overhead cost in numerical methods like finite element method (FEM).

- The automatic differentiation used in PINN computes accurate derivatives, unlike the numerical methods that include truncation errors.
- Transfer learning is the most powerful tool of all the data-based methods where once the model is trained with a set of parameters, it can quickly predict the results for another set of parameters with minimum computation cost. This is a useful characteristic of the model as an online soft sensor. The numerical model took the same computation time to solve the PDEs even if there is a small change in the single parameter.

PINN has been successfully applied across a wide range of research problems, including fluid mechanics (19–22), physics (23,24), biology (25,26), and supersonic compressible flow problems relevant to aerospace vehicles (27) and many more. Given the limitations of the original PINN in addressing certain complex problems, several recent modifications have been introduced to enhance its computational performance and solution accuracy. For example, the conservative PINN (cPINN) (28) uses a separate NN for each sub-domain within the computational domain to solve nonlinear conservation laws. Similarly, the eXtended PINN (29) employs a generalized decomposition method that can partition subdomains of any differential equation. This approach allows for the use of separate NNs in each sub-domain, facilitates efficient hyperparameter tuning, supports parallelization, and provides a greater representation capacity. Also, some studies have shown that using adaptive activation functions can improve the learning ability and convergence speed of NNs, especially when applied to solving forward and inverse differential equations using PINNs (30–32).

In the AM community, some researchers have used PINN models to study thermal history evolution in metal AM (33–36). Zhu et al. (37) predicted the domain temperature and melt pool dynamics using the PINN framework, comparing the learning efficiency and accuracy between "hard" and "soft" boundary condition cases. Their results showed that the PINN model could predict thermal history, melt pool velocity, and cooling rate with relatively less training data. Liao et al. (38) applied the PINN model to predict thermal history in the laser metal deposition (LMD) process. They trained the model with and without labeled training data, finding that the computational time to train the PINN model with partial temperature data at the top boundary was significantly reduced. They implemented the trained model to predict temperatures in the actual experiment. Although researchers have attempted to apply the PINN framework to the additive manufacturing process, the complex multiphysics problem of predicting thermal stress evolution during the laser scanning process has not yet been addressed. This is a critical area for future research in developing an ML-driven digital twin. Building a PINN model for thermal stress evolution and evaluating its performance against conventional numerical models is essential. It is important to note that while data-driven models (conventional ML models or PINNs) may initially require more time for training, once trained, they can easily be adapted to new parameter sets with minimal computational cost, making them superior to numerical models in this regard and they can be used for an online soft sensor.

To address the multiphysics problem of thermal stress evolution in the LMD process, this study focuses on developing a thermal-mechanical physics-informed neural network (PINN) model specifically designed to analyze the multiphysics problem of thermal stress evolution in an LMD process. The potential of PINN in addressing complex multiphysics problems offers a more efficient alternative to address the lasting challenges facing traditional numerical methods and data-driven ML methods. The paper is structured as follows: Section 2 describes the governing equations for the LMD process. In Section 3, different PINN architectures are compared, and the advantages of the current NN architecture for thermal and mechanical models are discussed.

Section 4 demonstrates the ability of the PINN model to predict the thermal field and stress without any labeled training data. Additionally, the acceleration of the training process using small simulation data is addressed. Finally, the model's transferability to different parameter sets is discussed. An outlook for the future direction of the work is presented in the last section.

## 2. Mathematical Formulation

This work focuses on the evolution of thermal stress during the deposition of a single layer of Ti-6Al-4V powder over the substrate of the same material at initial room temperature. Fig. 1 shows a schematic of the metal deposition process. A laser heat source starts scanning the substrate, creating a small melt pool where the powder particles melt into liquid and subsequently solidify as the laser moves to the next spot. Heat transfer occurs through three mechanisms: conduction, convection, and radiation. The primary focus of this study is to demonstrate the ability of the PIDL model to predict thermal stress evolution during laser scanning. Therefore, certain assumptions were made to simplify the study while still accurately representing the actual physics of the process. The following assumptions are made for this study:

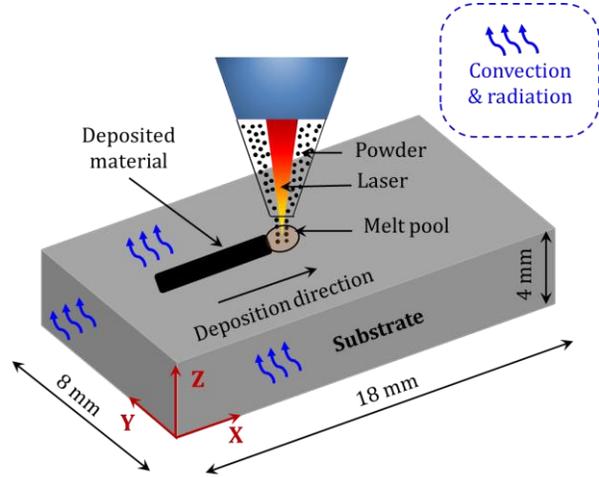

**Fig. 1:** Schematic of a LMD process.

- The deposited metal layer is thin relative to the substrate thickness, so it is assumed that the deposited mass does not significantly affect the overall temperature.
- The latent heat of fusion is disregarded.
- Melt pool fluid flow and evaporation are not taken into account.
- Both the substrate and the deposited material are considered homogeneous, with constant material properties.
- Only elastic thermal stresses are considered, while plastic deformation is neglected.

### 2.1 Governing equations

In this section, the governing equations defining the process are discussed. The energy equation for LMD can be defined as:

$$\frac{\partial(\rho C_p T)}{\partial t} = \kappa \left( \frac{\partial^2 T}{\partial x^2} + \frac{\partial^2 T}{\partial y^2} + \frac{\partial^2 T}{\partial z^2} \right) \quad (1)$$

where $\rho$ is the density of the material, $C_p$ is the heat capacity, $T$ is the temperature and $\kappa$ is the thermal conductivity. The thermal boundary conditions are given by:

$$-\kappa \frac{\partial T}{\partial n} = Q_{laser} + Q_{conv} + Q_{rad} \quad (2)$$

where $n$ is the normal to the surface, $Q_{laser}$ is the heat input by the laser heat source, $Q_{conv}$ is the convective heat loss, and $Q_{rad}$ is the radiative heat loss and given by:

$$Q_{laser} = -\frac{2\eta P}{\pi r_b^2} \exp\left(\frac{-2(x-vt)^2}{r_b^2}\right) \tag{3}$$

$$Q_{conv} = h(T - T_0) \tag{4}$$

$$Q_{rad} = \sigma\epsilon(T^4 - T_0^4) \tag{5}$$

where $\eta$ is the laser absorption coefficient, $P$ is the laser power, $r_b$ is the laser beam radius, $v$ is the laser scanning velocity, $h$ is the convective heat transfer coefficient, $\sigma$ is the Stefan-Boltzmann constant, $\epsilon$ is the emissivity, and $T_0$ is the ambient temperature. The bottom surface of the substrate has a finite temperature boundary condition. The initial temperature of the domain is equal to the ambient (298 K). For the mechanical analysis, the strain displacement relation can be given as:

$$\varepsilon_{xx} = \frac{\partial u}{\partial x} - \alpha(T - T_{ref}) \tag{6}$$

$$\varepsilon_{yy} = \frac{\partial v}{\partial y} - \alpha(T - T_{ref}) \tag{7}$$

$$\varepsilon_{zz} = \frac{\partial w}{\partial z} - \alpha(T - T_{ref}) \tag{8}$$

$$\varepsilon_{xy} = \frac{1}{2}\left(\frac{\partial u}{\partial y} + \frac{\partial v}{\partial x}\right) \tag{9}$$

$$\varepsilon_{yz} = \frac{1}{2}\left(\frac{\partial v}{\partial z} + \frac{\partial w}{\partial y}\right) \tag{10}$$

$$\varepsilon_{zx} = \frac{1}{2}\left(\frac{\partial u}{\partial z} + \frac{\partial w}{\partial x}\right) \tag{11}$$

where $\varepsilon_{ij}$ represents the strain tensor that depends on the displacement components $u$, $v$, and $w$ in the $x$, $y$, and $z$ directions respectively, $\alpha$ is the thermal expansion coefficient. Here, $u$, $v$, and $w$ are the independent field variables whose distribution over the domain needs to be identified over time. The stress-strain constitutive law can be given as:

$$\sigma_{xx} = \frac{E}{(1+v)(1-2v)}\left[\varepsilon_{xx}(1-v) + v(\varepsilon_{yy} + \varepsilon_{zz})\right] \tag{12}$$

$$\sigma_{yy} = \frac{E}{(1+v)(1-2v)}\left[\varepsilon_{yy}(1-v) + v(\varepsilon_{xx} + \varepsilon_{zz})\right] \tag{13}$$

$$\sigma_{zz} = \frac{E}{(1+v)(1-2v)}\left[\varepsilon_{zz}(1-v) + v(\varepsilon_{xx} + \varepsilon_{yy})\right] \tag{14}$$

$$\sigma_{xy} = \frac{E}{(1+v)}\varepsilon_{xy} \tag{15}$$

$$\sigma_{yz} = \frac{E}{(1+v)}\varepsilon_{yz} \qquad (16)$$

$$\sigma_{zx} = \frac{E}{(1+v)}\varepsilon_{zx} \qquad (17)$$

where $\sigma_{ij}$ represents the stress tensor components that depends on the Young's modulus, $E$ and Poisson's ratio $v$. The equilibrium equation can be given as:

$$\rho\frac{\partial^2 u}{\partial t^2} - \frac{\partial \sigma_{xx}}{\partial x} - \frac{\partial \sigma_{xy}}{\partial y} - \frac{\partial \sigma_{zx}}{\partial z} = 0 \qquad (18)$$

$$\rho\frac{\partial^2 v}{\partial t^2} - \frac{\partial \sigma_{xy}}{\partial x} - \frac{\partial \sigma_{yy}}{\partial y} - \frac{\partial \sigma_{yz}}{\partial z} = 0 \qquad (19)$$

$$\rho\frac{\partial^2 w}{\partial t^2} - \frac{\partial \sigma_{zx}}{\partial x} - \frac{\partial \sigma_{yz}}{\partial y} - \frac{\partial \sigma_{zz}}{\partial z} = 0 \qquad (20)$$

For the boundary conditions, the bottom boundary is treated as a fixed boundary, while all other boundaries are considered stress-free. The different process parameters and material properties used in the study are tabulated as:

Table 1. Process parameters and material properties of Ti6Al4V (39).

| Parameter | Value |
| --- | --- |
| Laser power (W) | 100 |
| Laser absorption coefficient | 0.4 |
| Laser beam radius (mm) | 1.5 |
| Laser scanning speed (mm/s) | 10 |

| Material property | Value |
| --- | --- |
| Density (kg/m$^3$) | 4122 |
| Heat capacity (J/kgK) | 831 |
| Thermal conductivity (W/mK) | 35 |
| Emissivity | 0.4 |
| Youngs Modulus (GPa) | 209 |
| Poisson ratio | 0.28 |

*2.2 Numerical modeling*

To validate the results predicted by the thermoelastic PINN model, FEM simulation data was used as the benchmark. This simulation data was later utilized to accelerate the training of the PINN model as well. The FEM simulations were conducted using the commercial software COMSOL Multiphysics®. The substrate dimensions were set to 18 mm × 8 mm × 4 mm, with a laser heat source of 100 W power scanning the substrate at a speed of 10 mm/s, as depicted in Fig.

1. The material properties and other process parameters are tabulated in Tables 1 and 2. The model employed a tetrahedral mesh consisting of 27,144 elements and 31,680 nodes. A non-uniform mesh was used, with a higher density of elements near the top surface and fewer elements toward the bottom. A 1-second laser scanning process was simulated. Temperature, displacement, and stress data were extracted at a frequency of 10 Hz for validation purposes. This simulation setup ensured a high-resolution dataset for accurate validation of the PINN model, particularly in capturing the thermal gradients and stress distribution throughout the substrate. The simulation took approximately 4 hours on an Intel Core i7 CPU using a single core.

## 3. PINN Methodology

PINN is one form of NNs that utilize the governing equations to either predict the results or accelerate the training process. They are particularly effective for modeling complex physical phenomena, such as fluid dynamics, heat transfer, and structural mechanics. Let's consider a non-linear PDE of a general form:

$$y_t + \mathcal{N}[y] = 0, \quad x \in \Omega, \quad t \in [0,T] \tag{21}$$

$$e := y_t + \mathcal{N}[y] \tag{22}$$

where $y(x,t)$ is a solution, $\mathcal{N}[\cdot]$ is a non-linear differential operator and $\Omega$ is a subset of $\mathbb{R}^D$. The initial conditions and boundary conditions are known and can be given as:

$$y(x,t) = y_{BC}, \quad x \in \partial\Omega \tag{23}$$

$$y(x,0) = y_{IC}, \quad x \in \Omega \tag{24}$$

where $\partial\Omega$ is the boundary. If this problem is solved using a fully connected NN, it will take the spatiotemporal resolution $(x,t)$ as the input and predict $y(x,t)$ at each iteration, calculating the data loss, given by Eq. 25, at the collocation points in the spatiotemporal domain. The advantage of PINN is that it includes three additional loss terms besides the data loss. The loss terms in PINN are:

$$\mathcal{L}_{Data} = \frac{1}{N}\sum_{n=1}^{N}|y_{pred}^n - y_{exact}^n|^2 \tag{25}$$

$$\mathcal{L}_{PDE} = \frac{1}{M}\sum_{m=1}^{M}|e(t^m, x^m, y^m)|^2 \tag{26}$$

$$\mathcal{L}_{BC} = \frac{1}{P}\sum_{p=1}^{P}|y_{BC,pred}^p - y_{BC,exact}^p|^2 \tag{27}$$

$$\mathcal{L}_{IC} = \frac{1}{Q}\sum_{q=1}^{Q}|y_{IC,pred}^q - y_{IC,exact}^q|^2 \tag{28}$$

$$\mathcal{L}_{Total} = w_1.\mathcal{L}_{Data} + w_2.\mathcal{L}_{PDE} + w_3.\mathcal{L}_{BC} + w_4.\mathcal{L}_{IC} \qquad (29)$$

Here *N, M, P,* and *Q* are the sampling points for each loss term. The number of sampling points for each loss term may vary. $\mathcal{L}_{Data}$ helps to learn the model from the labelled data, while $\mathcal{L}_{PDE}$, $\mathcal{L}_{BC}$, and $\mathcal{L}_{IC}$ accelerate training by penalizing the model when predicted values fail to satisfy the governing equations, boundary conditions, and initial conditions, respectively. In PINN models, data loss is optional. It helps in predicting more accurate results and accelerates the training. The total loss is the weighted average of these four loss terms where weights are assigned to different loss terms to balance the gradients, accelerate the convergence rate, and enhance the quality of final solutions. The current study assumes $\{1, 1, 1, 1e^{-4}\}$ for $w_1, w_2, w_3$, and $w_4$. The current PINN model was implemented using PyTorch and derivatives are calculated by the automatic differentiation (AD) technique (40). This technique is different and superior as compared to numerical differentiation like Taylor's series. AD uses the chain rule to calculate the derivative which is accurate up to the machine's precision. Finally, to measure the accuracy of the predictions, a relative *L₂* error is calculated between the predicted quantity *p* and exact function *f* as given by:

$$\mathcal{E}(p,f) = \left(\frac{1}{N}\sum_{i=1}^{N}[p(x_i) - f(x_i)]^2\right) \qquad (30)$$

where $\{x_i = 1, 2, ..., N\}$ are the collocation points scattered in the whole domain.

*3.1 Thermoelastic PINN*

In this study, three different PINN architectures were evaluated before finalizing the most efficient thermoelastic PINN. All three networks accurately predicted temperatures and stresses but differed in computational time. Network parameters were randomly initialized using the Glorot method, and the models were trained with the Adam optimizer, employing a learning rate of 2e-4 and the tanh activation function. The input to the network is scaled between -1 and 1. The output layer of the temperature network uses the Soft-plus activation function to ensure positive temperature values, while the output layer of the stress-displacement network is linear, with no activation function applied.

The first architecture consists of a single NN with 10 hidden layers and 64 neurons per layer. The input includes spatiotemporal coordinates *(x, y, z, t)*, and the output consists of temperature and displacement *(T, u, v, w)*. Stress components were predicted across the domain using displacement values and equations (4-18). The PIDL model required ~200,000 epochs, taking approximately 11 hours on a single Nvidia RTX A6000 GPU.

The second architecture features two separate NNs: one for temperature prediction and the other for displacement. The temperature network has 3 hidden layers with 64 neurons per layer, while the displacement network has 10 layers with 64 neurons per layer. This configuration was refined through trial and error. Both networks receive the same spatiotemporal inputs *(x, y, z, t)*. The temperature network predicts the temperature *T*, and the displacement network predicts the displacement components *(u, v, w)*. Initially, the temperature is predicted across the entire domain, and once the temperature network converges, its output is used as input to the displacement network to solve equations (4-6). The temperature network converged significantly faster, requiring only 25,000 epochs (approximately 21 minutes on the Nvidia RTX A6000 GPU), while the displacement network took ~200,000 epochs (approximately 8.2 hours). The key advantage of this architecture is that it avoids solving the energy equation for the full number of epochs (~200,000), which is primarily required for displacement field convergence.

The third and most efficient architecture also utilizes two NNs: one for temperature and another for combined stress and displacement prediction. The temperature network retains the same architecture as in the second configuration, but the displacement network is replaced by a stress-displacement network. The input includes spatiotemporal coordinates *(x, y, z, t)*, and the output consists of three displacement components and six stress tensor components. The advantage of this architecture is that, unlike the second architecture, which must compute stress at each iteration using equations (6-17), this model directly predicts the stress components in the output layer. This reduces the computational cost by avoiding the calculation of stress components in each iteration from strain-displacement and stress-strain relations, especially at the boundary points since the five boundaries have stress-free conditions. The stress-displacement network required ~200,000 epochs, taking approximately 6.5 hours on the same GPU for training without labeled data. Figure 2 presents the architecture of the fully connected thermoelastic PINN used in this study.

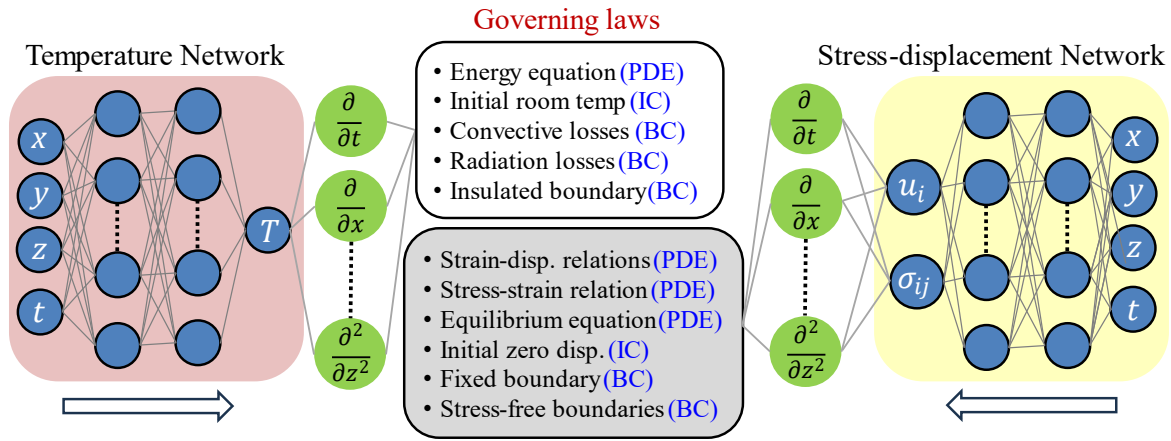

**Fig. 2:** Architecture of temperature network (red) and stress-displacement network (yellow) with the set of governing laws used in the loss functions.

*3.2 Collocation points*

Meshing plays a critical role in numerical modeling approaches like FEM, and similarly, the selection of collocation points is crucial in PINN models. A smaller number of collocation points may fail to capture the underlying physics, particularly in regions with steep gradients (41). On the other hand, using too many collocation points can significantly increase computational time. In this study, a non-homogeneous collocation point density was selected based on the physics of the problem. Since laser scanning is very fast, its effect is concentrated within only a few layers, resulting in a high-temperature gradient near the laser heat source, while the temperature remains relatively constant farther from the source. Therefore, a non-uniform collocation point distribution was chosen, as shown in Fig. 3, with a higher point density near the top layer and fewer points toward the bottom. Additionally, more collocation points were added around the laser center within a 2×2 mm area (indicated by red points in Fig. 3) to capture the dynamic effects more accurately. Further details about the selection and distribution of collocation points can be found in Liao et al. (38).

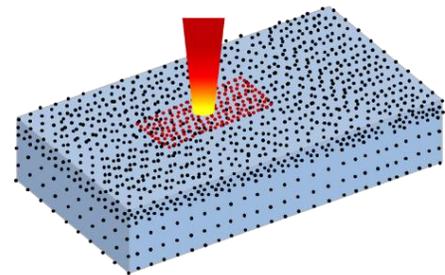

**Fig. 3:** Non-uniform distribution of collocation points.

## 4. Results and Discussions

To demonstrate the effectiveness of the thermoelastic PINN, three different cases of the forward problem are discussed in the following sub-sections. First, the ability of the thermoelastic PIDL to predict temperature, displacement, and stress fields without any labeled training data is evaluated. Second, the acceleration of the training process using small simulation data is examined. Third, the capability of the thermoelastic PINN to produce faster predictions when initialized with weights from a previously trained model, rather than random initialization, is explored. These results demonstrate that PINNs can serve as a surrogate model for studying thermal stress evolution. In all cases, FEM results are used as the benchmark for validation.

*4.1 Prediction without training data*

The thermoelastic PINN model can use the governing PDEs, initial conditions, and boundary conditions only to predict the temperature, displacement, and stress field. It is worth noting that no labeled training data is used to train the model. The temperature network took 26 minutes to train on a single Nvidia RTX A6000 GPU. Fig. 4, shows the comparison between the PINN's predicted temperature with the FEA temperature at time = 0.5 seconds. The results show that the PINN model can accurately predict the temperature field, with a root mean square error (RMSE) of 2.03 K. The selection of collocation points near the laser center is critical; insufficient points in this region can lead to inaccurate predictions of the high gradient around the laser center. Once the temperature field is predicted, it can be used in equations (6-8) to calculate the displacement field due to thermal expansion. The evolution of the different loss terms and the mean squared error (MSE) validation error is shown in Fig. 5. Initially, the IC loss has a higher value, and its contribution is balanced by assigning it a lower weight in the total loss (Eq. 29). The PDE loss starts off very low, then increases up to 16,000 iterations before decreasing again. This behavior is likely because, at the start, the network predicts a nearly uniform temperature across the domain, which easily satisfies Eq. (1) and results in a small residual. At this stage, however, the initial conditions (IC) and boundary conditions (BCs) are not yet satisfied, leading to a higher total loss. As the PINN model begins to capture the temperature gradient, the PDE residual increases resulting in higher PDE loss, and eventually, as the correct temperature distribution is learned, the PDE loss decreases again.

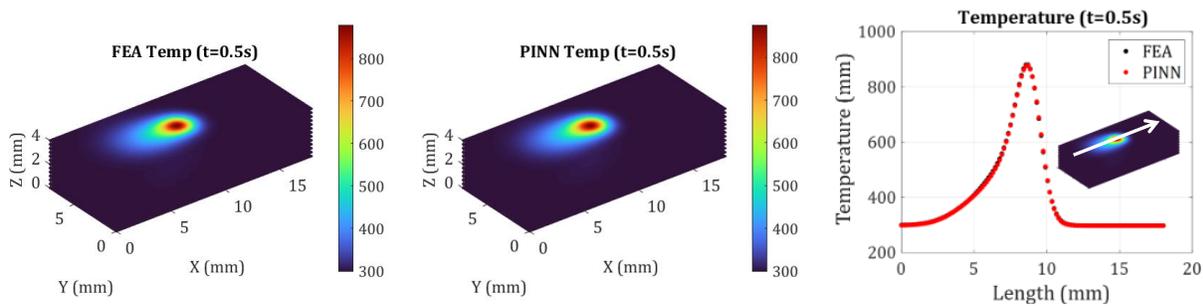

**Fig. 4:** Comparison of temperature field (a) FEA (b) PINN (c) top-surface along the centerline.

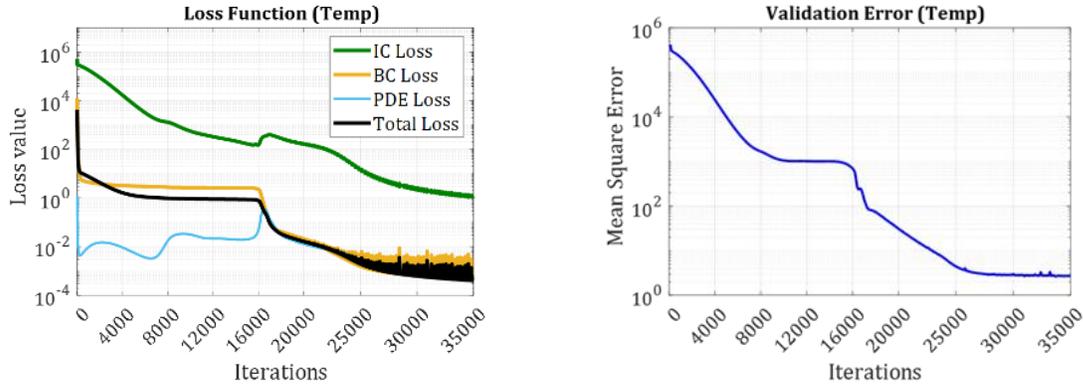

**Fig. 5:** (a) Loss function (b) Validation error evolution for the temperature network.

The predicted displacement is compared with the FEA results in Fig. 6, showing that the thermoelastic PINN model can accurately predict the displacement field. As indicated by Eq. (6-20), displacement is the only independent variable for the solid mechanics model and both strain and stress fields can be derived from it. Though, in this model, the stress tensor is also one of the outputs from the stress-displacement network, which helps to reduce the computational cost. If the model can accurately predict the displacement field, it will also predict the stress field accurately. Fig. 7 compares the predicted stress components with those from the FEA. The PINN model predicts the maximum values of $\sigma_x$ and $\sigma_y$ slightly lower than the FEA results. It is worth noting that in this model, melt pool dynamics are not considered. In actual cases, there is no stress at the laser center due to melt pool formation. Therefore, the critical areas are around the melt pool, where the PIDL model produces accurate predictions.

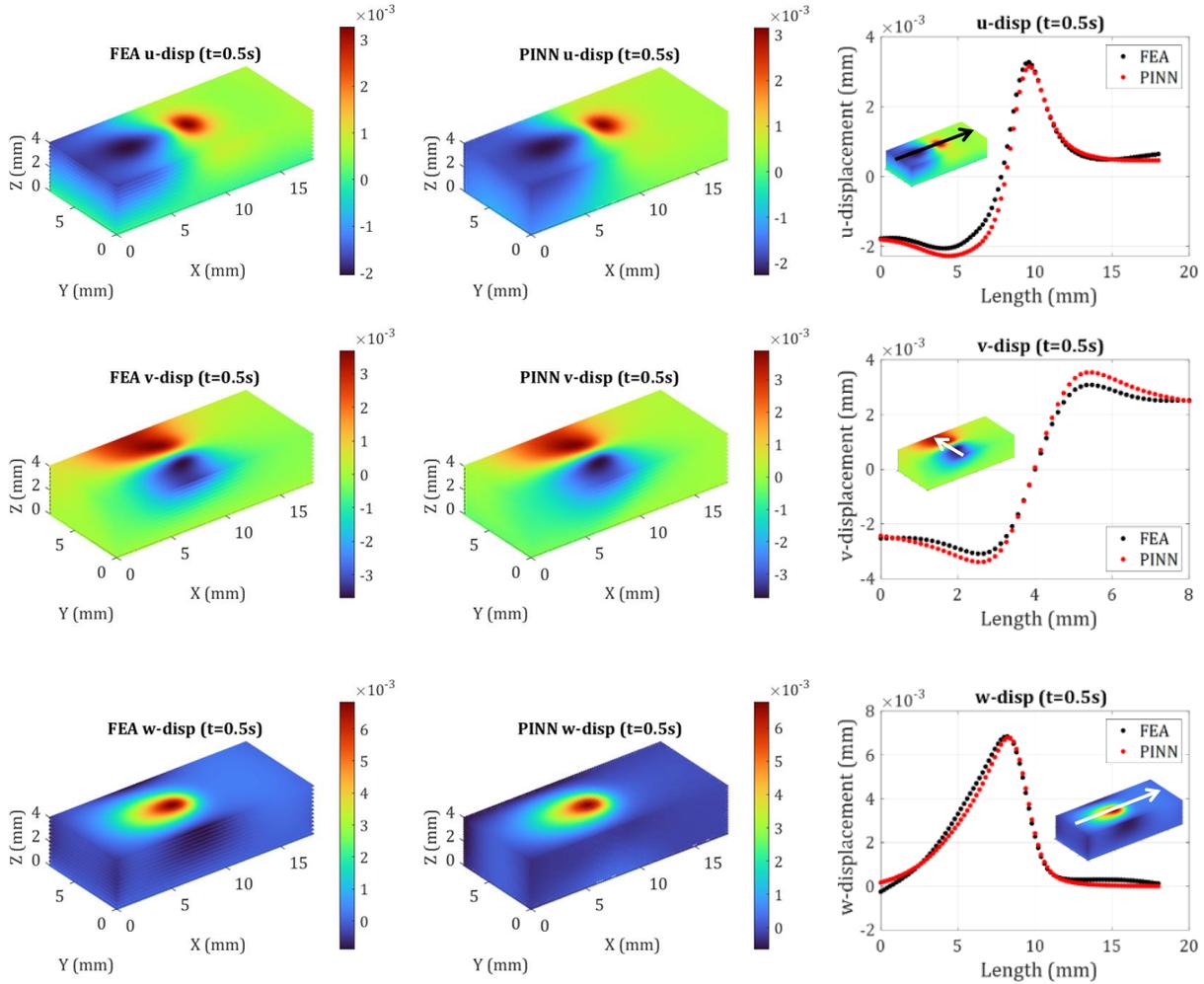

**Fig. 6:** Comparison of displacement field (a) FEA (b) PINN (c) top-surface along the centerline.

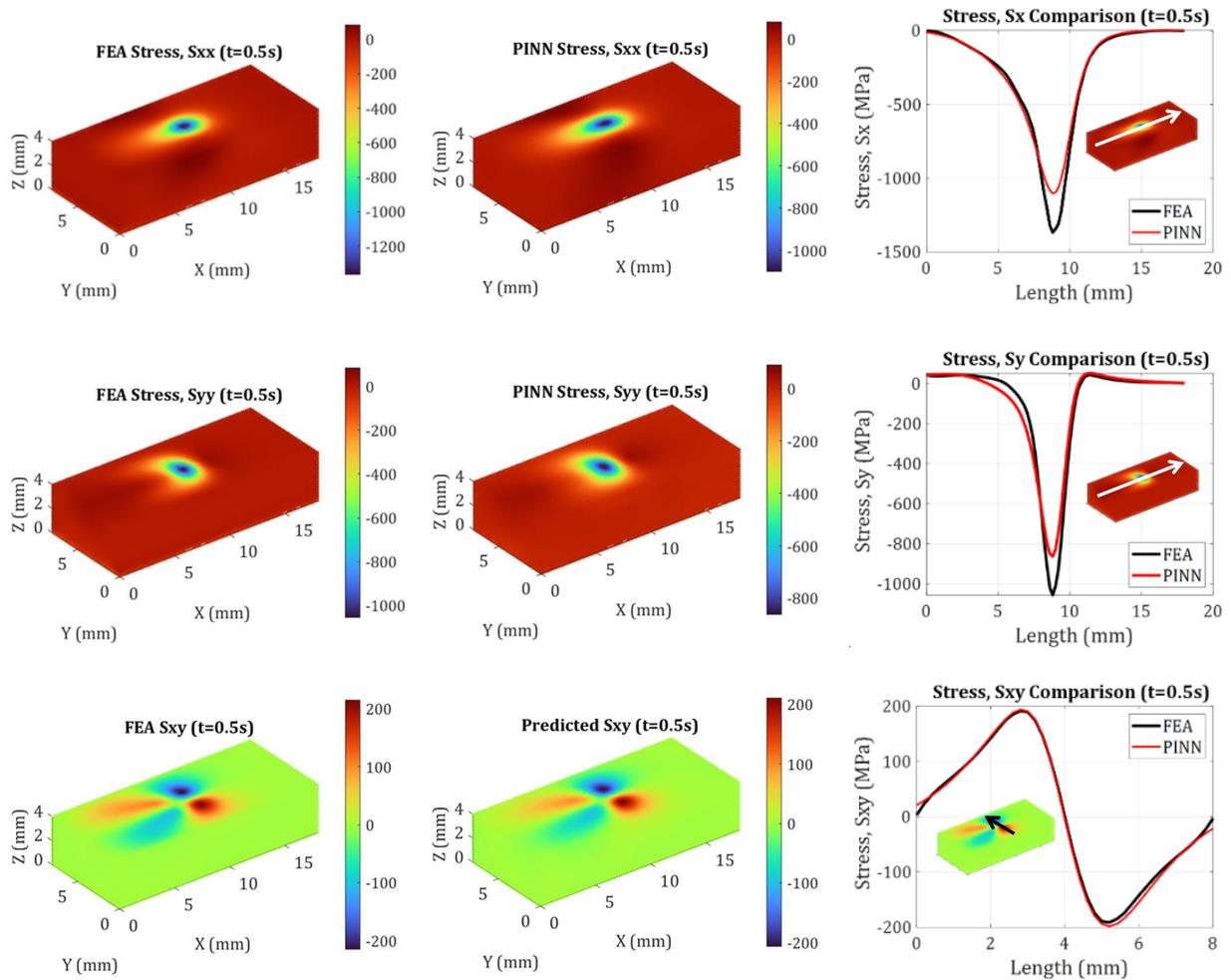

**Fig. 7:** Comparison of displacement field (a) FEA (b) PINN (c) top-surface along the centerline.

Fig. 8 illustrates the evolution of the different loss terms along with the MSE validation error. It is observed that the validation error decreases very slowly after 60,000 iterations. At this point, the PINN model can predict results, but with some differences in magnitude. To improve prediction accuracy, the model needs to be trained for 200,000 iterations. This process can be accelerated by using partial domain data, which is discussed in detail in the following subsection.

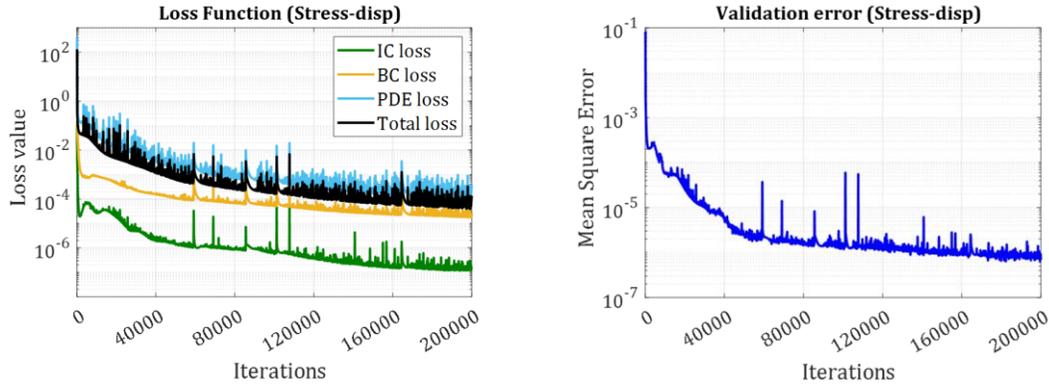

**Fig. 8:** (a) Loss function (b) Validation error evolution for the stress-displacement network.

*4.2 Prediction with training data*

In this section, the effect of using partial simulation data as a labeled training dataset is studied to improve accuracy and training efficiency. It is important to note that physical laws and labeled data are not enforced as "hard" constraints in the PINN model, meaning the model is trained to minimize the loss rather than reduce it to zero. The advantage of this approach is that simulation or experimental data can be easily incorporated without over-constraining the model. Additionally, there is no minimum required dataset size for the model to converge.

In this study, displacement and stress simulation data, consisting of 78,000 points over the time range from 0 to 0.7 seconds, were used to train the stress-displacement network. In Fig. 9, the predicted stress component ($\sigma_x$) was compared for models trained with and without labeled data for 70000 iterations. It took approximately 2.5 hours to get the model trained. It was observed that the model trained with labeled data converged significantly faster than the model without labeled data with relatively low validation error as shown in Fig. 9.

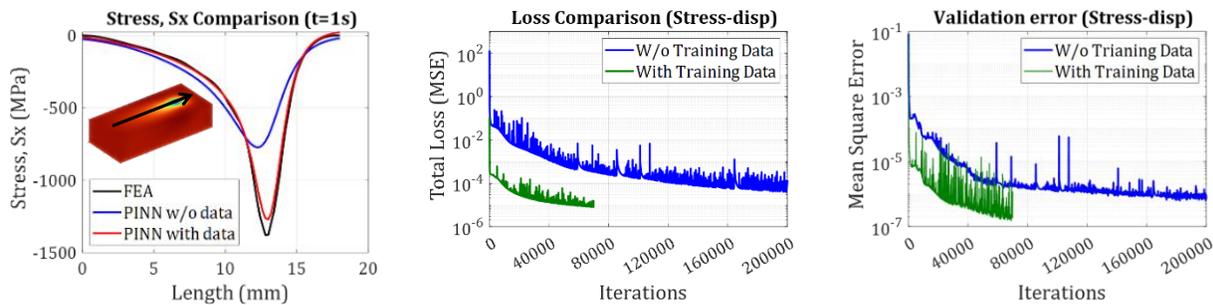

**Fig. 9:** Fast and accurate prediction of PINN model using partial simulation data witnessed by (a) stress at top-surface and along the centerline (b) Total loss (c) validation error comparison.

*4.3 Prediction using pre-trained model*

The primary advantage of using data-driven ML models over conventional numerical models is their transferability. Once trained on a specific set of parameters, an ML model can predict results for different parameter sets with minimal computational cost. (42). This characteristic makes data-driven models highly suitable for soft sensing applications. Fig. 10 shows the evolution of the loss function, temperature, and stress field for a laser power of 150 W and a scanning speed of 5 mm/s. The temperature and stress-displacement networks converged after 1500 iterations, which took approximately 0.8 minutes and 2.6 minutes, respectively—significantly less computational time compared to the model training phase. The results closely align with benchmark FEA data, demonstrating the transferability and accuracy of the PINN model.

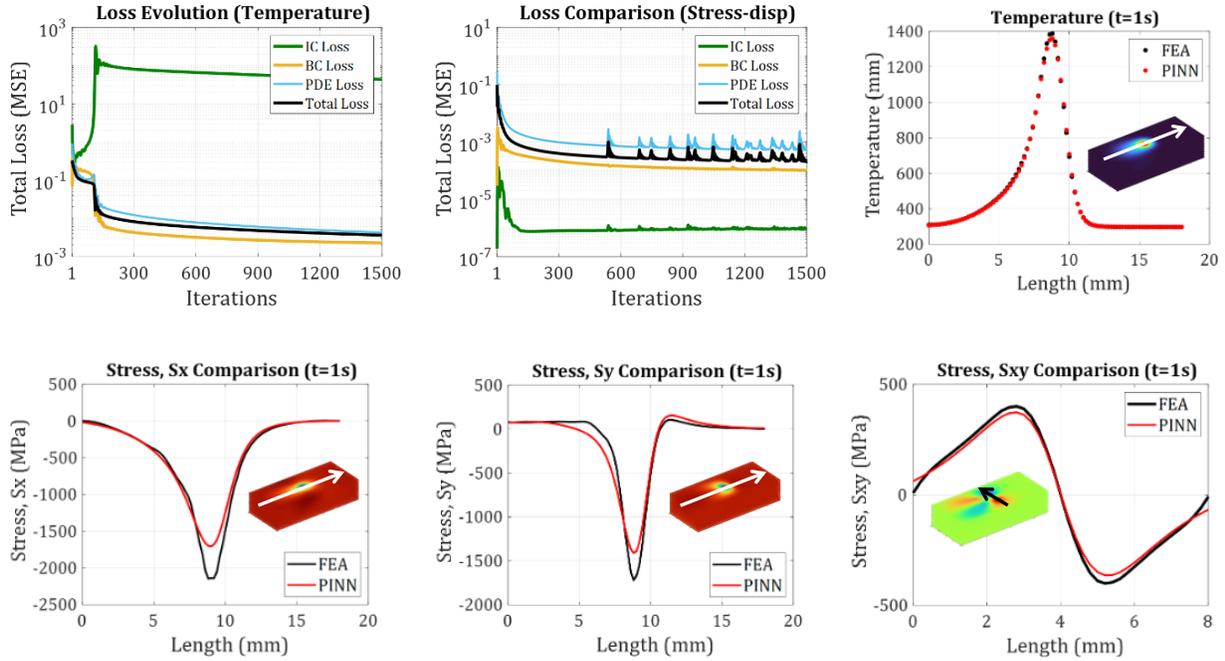

**Fig. 10:** Loss function evolution of (a) Temperature network (b) Stress-displacement network. On top surface along the center line at t = 1 second (c) Temperature (d) Stress, $S_x$ (e) Stress, $S_y$ (f) Stress, $S_{xy}$

## 4. Conclusions and Outlook

This study presents a PINN approach to predict temperature and thermal stress evolution in the laser metal additive manufacturing (AM) process. A thermoelastic PINN model is developed to predict stress evolution in LMD. This paper demonstrates the application of scientific machine learning (SciML) as an alternative to conventional physics-based models for efficiently solving complex multiphysics problems, particularly in parametric studies. The major conclusions of this work are as follows:

- PINN can be used as an alternative method to facilitate the prediction the temperature and thermal stress evolution in the LMD process without requiring labeled training data. Therefore, the thermoelastic PINN model can be used as a soft sensor for fast prediction of thermal stress evolution.
- The most efficient PINN architecture involves two separate networks for temperature and stress field predictions. The stress-displacement network outputs the three displacement and six stress components, avoiding extra calculations and improving computational efficiency.
- Incorporating small, labeled data from simulations makes the PINN model more accurate and training efficient.
- The thermoelastic PINN model can predict results for different sets of parameters very quickly (~3 minutes), compared to physics-based simulation models, demonstrating its transferability and suitability for real-time predictions.

This study represents the first application of the PINN approach to predict thermal stress evolution in the LMD process. Although PINN is highly efficient for parametric studies, it is less efficient for a single set of laser parameters, where physics-based simulations may perform better. The authors believe there is potential to optimize the NN architecture further, which could reduce computation time. This optimization could be explored in future work. Additionally, the model can be expanded to include more physics, such as phase changes, temperature-dependent material properties, and plastic deformation. The PINN model can also be applied to inverse problems, such as learning equation parameters like the laser absorption coefficient or material properties.

**Acknowledgments**
The authors would like to thank the financial support of the National Science Foundation under the grant CMMI- 2152908.

**Data availability**
Data will be made available on request.